%% file: acl2026.tex
\documentclass[11pt]{article}

\usepackage[preprint]{acl}

\usepackage{times}
\usepackage{latexsym}

\usepackage[T1]{fontenc}

\usepackage[utf8]{inputenc}

\usepackage{microtype}

\usepackage{inconsolata}

\usepackage{graphicx}

\usepackage{amsfonts}
\usepackage{amsmath}

\usepackage{multirow}
\usepackage{subcaption}
\usepackage{booktabs}

\usepackage{arydshln}
\newcommand{\mycdashline}[1]{%
    \noalign{\vskip 2pt}  
    \cdashline{#1}        
    \noalign{\vskip 2pt}  
}

\input{isabelle}

\title{Beyond Gold Standards: Epistemic Ensemble of \\LLM Judges for Formal Mathematical Reasoning}

\author{Lan Zhang\textsuperscript{1},
  Marco Valentino\textsuperscript{2},
  Jordan Meadows\textsuperscript{3},
  Andr\'e Freitas\textsuperscript{1,3}\\
  \textsuperscript{1}Department of Computer Science, University of Manchester, United Kingdom\\
  \textsuperscript{2}School of Computer Science, University of Sheffield, United Kingdom\\
  \textsuperscript{3}Idiap Research Institute, Switzerland\\
  \texttt{lan.zhang-6@postgrad.manchester.ac.uk}\\
  \texttt{m.valentino@sheffield.ac.uk}\quad\texttt{\{jmeadows,andre.freitas\}@idiap.ch}
}

\begin{document}
\maketitle
\begin{abstract}
Statement autoformalization plays a crucial role in formal mathematical reasoning. However, while large language models (LLMs) have shown promise, automatically evaluating the formalization of mathematical statements remains a challenge. Existing LLM-as-a-judge approaches typically use coarse-grained, generic evaluation criteria, with limited effectiveness for advanced formal reasoning, where quality depends on nuanced, multi-granular dimensions. In this work, we take a step toward addressing this gap by introducing a new multi-dimensional framework for autoformalization evaluation. The proposed method is based on an epistemically and formally grounded (EFG) ensemble of LLM judges, assessing logical preservation (LP), mathematical consistency (MC), formal quality (FQ), and formal validity (FV) through fine-grained atomic properties and theorem-prover verification. Experiments on miniF2F and ProofNet show that fine-grained evaluation outperforms coarse-grained judgments, achieving the highest pairwise agreement with human rankings in 17 out of 20 comparisons. Notably, fine-grained evaluation improves GPT-4.1-mini's pairwise agreement with human judgments by up to 12.7 percentage points, reaching levels comparable to human inter-annotator agreement. These findings demonstrate that LLM judges, especially when guided by a well-defined set of evaluation principles, can offer scalable, interpretable, and reliable support for formal mathematics.\footnote{Code and datasets are available at \url{https://github.com/lanzhang128/autoformalization_evaluation}}
\end{abstract}

\begin{figure*}[!t]
    \centering
    \includegraphics[width=\textwidth]{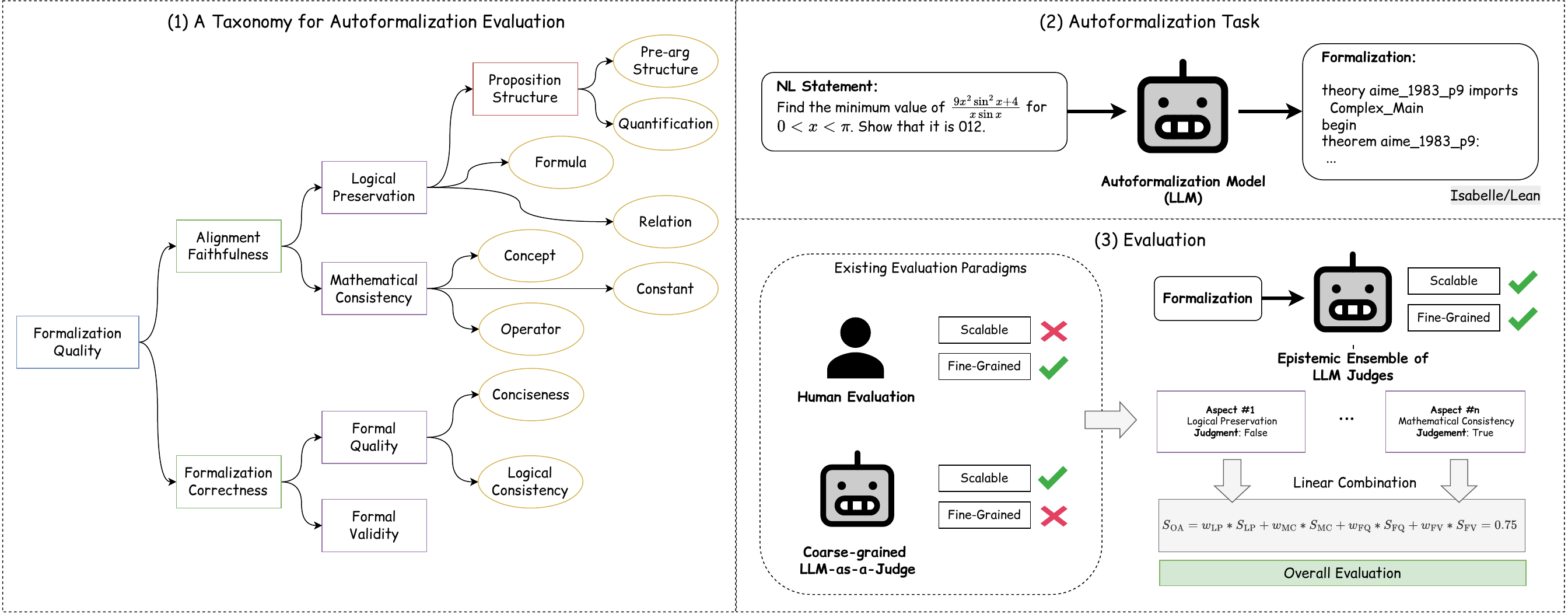}
    \caption{Overview of the proposed framework for automatically evaluating autoformalization. Our method leverages ensembles of LLM judges based on epistemic criteria organized into a taxonomy, while also providing explanations for their judgments. A linear evaluation model aggregates the criteria into a single overall assessment to serve as a proxy for human evaluation.}
    \label{fig:approach}
\end{figure*}

\section{Introduction}\label{sec:intro}
With the increasing capabilities of large language models (LLMs), reasoning in formal mathematics has received growing attention in recent years~\citep{jiang2023draft,xin2025deepseekproverv, lin2025goedelproverfrontiermodelopensource, ahuja2025improver,poesia2024learning, yu2025formalmathbenchmarkingformalmathematical, yang2024formalmathematicalreasoningnew}. Autoformalization, which serves as a bridge between natural language and formal language domains, plays a critical role in improving accessibility to formal verification tools for formal reasoning \citep{pan2023logic,quan2025faithful,quan-etal-2024-verification}. Despite recent advances in autoformalization with LLMs~\citep{wu2022autoformalization, zhang-etal-2024-consistent}, its evaluation remains a long-standing challenge. While syntactic correctness~\citep{zhang-etal-2024-consistent,zhang-etal-2025-autoformalization,zhang-etal-2025-masa} can be rigorously verified using theorem provers (TPs) such as Isabelle~\citep{paulson2000isabelle} or Lean~\citep{lean}, assessing semantic alignment or consistency~\citep{zhang-etal-2024-consistent} between natural and formal language statements is still a significant open problem. 

While human experts can provide direct assessments, formalizations in highly specialized domains are extremely time-consuming to evaluate and prone to errors. Human evaluation becomes particularly challenging when dealing with increasingly complex and novel mathematical objects~\citep{zhang-etal-2025-autoformalization}, where the required background knowledge grows substantially. This creates a clear scalability bottleneck for autoformalization evaluation.

On the other hand, although reference-based metrics~\citep{li2024autoformalize,poiroux-etal-2025-reliable,liu2025rethinking} enable automatic evaluation, they introduce several limitations: (i) they require ground-truth formalizations to be correct, which has already been identified as an issue in formal benchmarks such as ProofNet~\citep{poiroux-etal-2025-reliable}; (ii) they rely on models to infer equivalence, which may introduce errors that are unrelated to the quality of the formalization itself; and (iii) they depend heavily on a formal execution environment and thereby are constrained by syntactic correctness, even though syntactically incorrect code may still be semantically meaningful and useful for refinement. These limitations highlight the need for reference-free evaluation methods for autoformalization.

In this paper, we take a step toward addressing this gap by introducing a systematic reference-free method for evaluating autoformalization for formal mathematical reasoning. The proposed method is based on an \emph{epistemically and formally grounded (EFG)} ensemble\footnote{The "ensemble" term used in this paper refers to using the same backend LLM to assess different aspects, unlike setups with multiple backend LLMs judging the same aspect. While more diverse models for different aspects could enhance generalizability, our results confirm that a single-backend ensemble suffices for reliable evaluation.} of LLM Judges~\citep{zheng2023judging}, grounded on a taxonomy for multi-dimensional autoformalization evaluation. We break down the components involved within the context of autoformalization tasks and organize them into a taxonomy composed of four core aspects: \textit{logical preservation} (LP), \textit{mathematical consistency} (MC), \textit{formal quality} (FQ), and \textit{formal validity} (FV), along with a set of atomic properties that serve as computable estimators of these features. We hypothesize that such properties can serve as an effective proxy for human evaluators in mathematical autoformalization settings. Specifically, in this framework (Figure~\ref{fig:approach}), LP, MC, and FQ are assessed by LLM judges, while FV is verified using theorem provers. A linear evaluation model integrates the estimators into a \textit{unified}, \textit{interpretable} evaluation outcome, enabling \textit{scalable}, \textit{multi-dimensional}, and \textit{automated} assessment.

\begin{table*}[t]
  \centering
  \small
  \begin{tabular}{p{0.27\textwidth} p{0.36\textwidth} p{0.27\textwidth}}
    \toprule
\textbf{Case 1:} miniF2F Isabelle\newline
\textit{Unpreserved Logic}\newline
Find the minimum value of $\frac{9x^2\sin^2 x + 4}{x\sin x}$ for $0 < x < \pi$. Show that it is 012.
    
\begin{lstlisting}[language=isabelle,aboveskip=0pt,belowskip=-3pt]
theorem aime_1983_p9:
  fixes x::real
  assumes "0<x" "x<pi"
  shows "12 \<le> ((9 * (x^2 * 
    (sin x)^2)) + 4) / (x * sin x)" 
    sorry
\end{lstlisting}
& 
\textbf{Case 2:} Proofnet Lean4\newline
\textit{Inconsistent Mathematics}\newline
Let $A$ be a nonempty set of real numbers which is bounded below. Let $-A$ be the set of all numbers $-x$, where $x \in A$. Prove that $\inf A=-\sup (-A)$.
    
\begin{lstlisting}[language=isabelle,aboveskip=0pt,belowskip=0pt]
theorem exercise_1_5 
  (A minus_A : Set ℝ) (hA : A.Nonempty) 
  (hA_bdd_below : BddBelow A) 
  (hminus_A : minus_A = {x | -x ∈ A}) :
  Inf A = Sup minus_A :=sorry
\end{lstlisting}
&
\textbf{Case 3:} miniF2F Lean4\newline
\textit{Redundant Hypothesis}\newline
Simplify $(2-2i)(5+5i)$, where $i^2 = -1.$ Show that it is 20.

\begin{lstlisting}[language=isabelle,,aboveskip=0pt,belowskip=0pt]
theorem mathd_algebra_110
  (q e : ℂ)
  (h₀ : q = 2 - 2 * Complex.I)
  (h₁ : e = 5 + 5 * Complex.I) :
  q * e = 20 := sorry
\end{lstlisting}\\
    \bottomrule
  \end{tabular}
\caption{Typical challenges involved in formalization evaluation.}
  \label{tab:prob}
\end{table*}

To validate the proposed framework, we conduct human evaluations of rankings on Lean4, using both human-written formalizations from existing datasets~\citep{zheng2022miniff, azerbayev2023proofnet} and LLM-generated formalizations. We further analyze how LLM judges and humans differ in the extent to which individual dimensions contribute to the final assessment when evaluating flawed formalizations on both Lean4 and Isabelle/HOL. Finally, we examine whether an ensemble of LLM judges (EFG) can reliably produce consistent and less biased overall evaluations, thereby supporting a fully automated evaluation protocol that minimizes human involvement while complementing human judgment.

Overall, our experiments demonstrate that fine-grained LLM judges are generally more effective than coarse-grained LLM judges and achieve ranking agreement comparable to that of human evaluators. Our contributions can be summarized as follows:

\begin{enumerate}
    \item We introduce an epistemically and formally grounded (EFG) taxonomy for formal mathematical reasoning, encompassing key dimensions such as logical preservation, mathematical consistency, formal quality, and formal validity. These dimensions are operationalized through a set of interpretable atomic properties serving as estimators.
    \item We develop a fine-grained and interpretable framework for automatic evaluation of autoformalization based on this taxonomy, which aggregates outputs from LLM judges specialized in the defined atomic properties via a linear ensemble.
    \item We investigate the impact of coarse-grained versus fine-grained evaluation designs using human evaluation. Our findings show that smaller LLMs guided by EFG criteria can outperform larger LLMs relying on coarse-grained judgments.
    \item We empirically validate the effectiveness of EFG ensembles for autoformalization evaluation. Results show strong agreement with human rankings, no evidence of cross-model-family bias, and robustness to sampling variability, indicating that EFG ensembles serve as reliable proxies for evaluating formalization quality.
\end{enumerate}

\section{Methodology}\label{sec:method}

\subsection{Formalization Evaluation}

Mathematical \textit{autoformalization} is the task of automatically transforming an informal mathematical statement $s$, expressed in natural language and LaTeX symbols from domain $\mathcal{S}$, into a formal representation $\phi$ within a formal language $\mathcal{F}$~\citep{zhang-etal-2024-consistent}. \textit{Formalization evaluation} refers to the systematic assessment of the correctness of a candidate formalization (produced either by a human or a model) and the assignment of an evaluation function (either a score or a binary function). The conventional evaluation approach relies on human judgment of correctness, often based on syntactic validity within a theorem prover, and typically results in a binary score (e.g., pass). However, such binary assessment \textbf{treats all incorrect formalizations equally} and fails to convey more detailed qualitative feedback.

Case studies~\citep{wu2022autoformalization} of autoformalizations produced by a particular model on a specific dataset can provide more nuanced insights, but they require human effort and tend to be tied to the particular combination of dataset, model, and formal language, limiting their generalizability to other settings. To this end, we aim to obtain an autoformalization evaluation function that has a twofold purpose: (i) serving as an automated proxy for human evaluators and (ii) providing a qualitative description around a set of formalization properties, while being agnostic towards any particular solver.

To illustrate the broader evaluation challenge, we present examples from human-validated, ground-truth formalizations in commonly used datasets in Table~\ref{tab:prob}. In Case 1, the original natural language statement explicitly asks for the minimum value, but the formalization merely asserts a lower bound for the expression, omitting the critical information that the minimum is actually attained for some value of $x$. This highlights a failure to preserve the underlying logical intent of the natural language. In Case 2, the formalization omits a negation on the right-hand side of an equality, contradicting the natural language statement and revealing a semantic inconsistency in the mathematical content. In Case 3, while the formalization is semantically correct, it introduces unnecessary variables and hypotheses, increasing complexity and redundancy, an undesirable property for concise and readable formalizations. Motivated by these observations, we propose a fine-grained taxonomy for formalization evaluation that systematically qualifies a set of target properties that should be satisfied.

\subsection{A Taxonomy for Formalization Evaluation}

We start by introducing a taxonomy which elicits the constituting parts entailed by a formalization evaluation framing  a set of well-defined computable properties (Figure~\ref{fig:approach}):

\noindent\textbf{Tier 1.} A correct candidate formalization $\phi$ of a natural language statement $s$ should satisfy two key properties:

\begin{itemize}
    \item \textbf{Alignment Faithfulness (AF):} As a representation of $s$ in the formal domain, $\phi$ must accurately capture the intended semantics of $s$ from the natural language domain.
    \item \textbf{Formalization Correctness (FC):} As a construct within the formal domain, $\phi$ must be valid, natural and well-formed within the conventions of the target formalism. 
\end{itemize}

\noindent\textbf{Tier 2 for AF.} To ensure that a formalization $\phi$ faithfully represents a source statement $s$, it is essential that $\phi$ maintains the logical structure and mathematical intent inherent in $s$. We introduce two conceptual properties to characterize this alignment:

\begin{itemize}
    \item \textbf{Logical Preservation (LP)} captures the degree to which the logical content and inferential structure of $s$ is preserved within $\phi$.
    \item \textbf{Mathematical Consistency (MC)} reflects how coherently and accurately the mathematical objects and operations present in $s$ are represented in $\phi$.
\end{itemize}

\noindent\textbf{Tier 2 for FC.} To clarify the intrinsic properties, which can be assessed independently of any source statement $s$, of a formalization $\phi$, we present the following conceptual properties:

\begin{itemize}
    \item \textbf{Formal Quality (FQ)} captures the structural clarity and conciseness of $\phi$ (e.g., non-redundant statements).
    \item \textbf{Formal Validity (FV)} refers to whether $\phi$ is syntactically valid within the target formal system. A formally valid expression conforms to all grammatical and structural rules of the formal language.
\end{itemize}

\noindent\textbf{Tier 3 - Operable Atomic Properties (OAPs).} The properties previously articulated are then materialized into a set of Operable Atomic Properties (OAPs), which are functions designed to be both interpretable and computable. We define a total of 9 such properties, each capturing a distinct and measurable aspect of a candidate formalization: Pre-arg Structure, Quantification, Formula, Relation, Concept, Constant, Operator, Conciseness, and Logical Consistency. The detailed description of each OAP is provided in Appendix~\ref{app:oap}.

An evaluation metric for autoformalization can be constructed as a linear combination of Tier-2 aspects:
\begin{multline*}
    S(s,\phi)=w_\mathrm{LP} S_\mathrm{LP}(s,\phi)+w_\mathrm{MC} S_\mathrm{MC}(s,\phi)\\
    +w_\mathrm{FQ} S_\mathrm{FQ}(\phi)+w_\mathrm{FV} S_\mathrm{FV}(\phi)
\end{multline*}
where $S_\mathrm{LP},S_\mathrm{MC},S_\mathrm{FQ},S_\mathrm{FV}\in[0,1]$, and the weights $w_\mathrm{LP},w_\mathrm{MC},w_\mathrm{FQ},w_\mathrm{FV}>0$ satisfy the constraint $w_\mathrm{LP}+w_\mathrm{MC}+w_\mathrm{FQ}+w_\mathrm{FV}=1$.

\section{Empirical Validation}

\subsection{Alignment Between LLM Judges and Human Rankings}
We first evaluate how well LLM judges with different settings can rank formalizations compared with human evaluation. To this end, we first randomly sample 50 natural language statements from the validation split of miniF2F~\citep{zheng2022miniff} and another 50 from ProofNet~\citep{azerbayev2023proofnet}. We then ask two human evaluators to independently rank~\footnote{Equivalence is allowed in the ranking process, as correct formalizations can be written in various ways, and even flawed formalizations may be difficult to distinguish in ranking.} formalizations from three sources: human-written ground truth, GPT-4.1~\citep{openai2024gpt4} zero-shot autoformalization, and Qwen2.5-7B~\citep{qwen2025qwen25technicalreport} few-shot autoformalization (8 exemplars), while also providing an overall assessment on a 5-point scale (criteria shown in Table~\ref{tab:anno} in Appendix~\ref{app:prompt}). We use two accuracy-based metrics, exact match (EM) and pairwise match (PM) of rankings, to measure agreement between the two human evaluators. The EM and PM scores are 38.00\% and 70.00\% on miniF2F, and 22.00\% and 68.67\% on ProofNet. From the human evaluations, we observe three findings: (1) even human evaluators can disagree substantially on the exact rankings of formalizations from multiple sources; (2) agreement on pairwise rankings is higher, though notable disagreement remains; and (3) human evaluators find it more difficult to agree on formalizations of more complex natural language statements (e.g., from ProofNet) than on formalizations of simpler statements (e.g., from miniF2F).

\begin{table*}[t]
  \centering
  \small
  \begin{tabular}{llcccccccc}
    \toprule
    & & \multicolumn{4}{c}{miniF2F} & \multicolumn{4}{c}{ProofNet}\\
    LLM & Setup & EM-H1 & EM-H2 & PM-H1 & PM-H2 & EM-H1 & EM-H2 & PM-H1  & PM-H2\\
    \midrule
    DeepSeek-Math-7B & OA-TP & 18.00 & 16.00 & 39.33 & 39.33 & 6.00 & \textbf{16.00} & 51.33 & 46.67\\
    DeepSeek-Math-7B & DA-T2 & 18.00 & \textbf{24.00} & 44.67 & 45.33 & 18.00 & 10.00 & 46.00 & 39.33\\
    DeepSeek-Math-7B & OAP & \textbf{20.00} & 16.00 & \textbf{47.33} & \textbf{49.33} & \textbf{20.00} & 8.00 & \textbf{60.00} & \textbf{48.00}\\
    \mycdashline{1-10}
    Qwen2.5-Coder-7B & OA-TP & 10.00 & 22.00 & 40.00 & 46.67 & 10.00 & 10.00 & 38.67 & 39.33\\
    Qwen2.5-Coder-7B & DA-T2 & \textbf{22.00} & \textbf{22.00} & \textbf{50.67} & \textbf{52.00} & 12.00 & 20.00 & 48.00 & 56.00\\
    Qwen2.5-Coder-7B & OAP & 16.00 & 12.00 & 44.67 & 46.67 & \textbf{14.00} & \textbf{20.00} & \textbf{56.67} & \textbf{62.00}\\
    \mycdashline{1-10}
    GPT-4.1-mini & OA-TP & 22.00 & 26.00 & 60.00 & 58.67 & 22.00 & \underline{\textbf{36.00}} & 65.33 & 68.00\\
    GPT-4.1-mini & DA-T2 & 16.00 & 24.00 & 57.33 & 56.67 & 28.00 & 24.00 & 66.00 & 66.00\\
    GPT-4.1-mini & OAP & \textbf{40.00} & \underline{\textbf{38.00}} & \underline{\textbf{72.67}} & \underline{\textbf{64.00}} & \textbf{30.00} & 26.00 & \textbf{69.33} & \textbf{69.33}\\
    \mycdashline{1-10}
    GPT-4.1 & OA-TP & 32.00 & 24.00 & 64.67 & 53.33 & 30.00 & 26.00 & 68.00 & 65.33\\
    GPT-4.1 & DA-T2 & 38.00 & \textbf{34.00} & 67.33 & 56.00 & 30.00 & 28.00 & 67.33 & 67.33\\
    GPT-4.1 & OAP & \underline{\textbf{40.00}} & 32.00 & \textbf{69.33} & \textbf{58.67} & \textbf{32.00} & \textbf{34.00} & \textbf{69.33} & \underline{\textbf{72.00}}\\
    \mycdashline{1-10}
    GPT-5.4 & OA-TP & 20.00 & 36.00 & 59.33 & 59.33 & 20.00 & 22.00 & 61.33 & 63.33\\
    GPT-5.4 & DA-T2 & \textbf{36.00} & \textbf{38.00} & 68.00 & \textbf{62.00} & 36.00 & 24.00 & 70.67 & 62.00\\
    GPT-5.4 & OAP & 34.00 & 32.00 & \textbf{69.33} & 60.00 & \underline{\textbf{36.00}} & \textbf{24.00} & \underline{\textbf{75.33}} & \textbf{70.67}\\
    \bottomrule
  \end{tabular}
  \caption{Results of using LLM judges to evaluate Lean4 formalizations from three sources: Human, GPT-4.1 and Qwen2.5-7B. The metrics include: (\textbf{EM}): the percentage of rankings that exactly match human rankings; (\textbf{PM}): the percentage of pairwise rankings that match human pairwise rankings. \textbf{H1} and \textbf{H2} denote two human annotators. The evaluation setups are: (\textbf{OA-TP}): overall assessment by an LLM judge with theorem prover feedback injected into the prompt; (\textbf{DA-T2}): directly assessing the Tier-2 aspects LP, MC, and FQ using LLM judges, and FV using the theorem prover, then averaging the results; (\textbf{OAP}): independently assessing the OAPs using LLM judges and synthesizing the Tier-2 aspects LP, MC, and FQ using a weighted average, where each OAP is weighted equally within each Tier-2 aspect, and FV using the theorem prover, then averaging the results. \textbf{Bold} represents the setup with the highest score within the same LLM and \underline{Underline} represents the setup with the highest score overall.}
  \label{tab:pair}
\end{table*}

\subsubsection{Results \& Discussions}
We study five LLMs as judges for autoformalization evaluation: DeepSeek-Math-7B~\citep{shao2024deepseekmath}, Qwen2.5-Coder-7B~\citep{qwen2025qwen25technicalreport}, GPT-4.1-mini, GPT-4.1, and GPT-5.4~\citep{openai2025gpt5}. We consider three settings: (OA-TP): LLM judges directly provide an overall assessment score based on the same criteria used in the human evaluation, together with detailed error feedback from the theorem prover; (DA-T2): LLM judges directly assess the Tier-2 aspects LP, MC, and FQ, while the theorem prover determines FV, and the resulting scores are averaged to obtain a final score between 0 and 1; and (OAP): LLM judges independently assess the OAPs corresponding to the Tier-2 aspects LP, MC, and FQ, after which the Tier-2 scores are synthesized through weighted averaging, with each OAP weighted equally within its corresponding Tier-2 aspect. FV and the final scores are then obtained in the same manner as in setting (ii). These scores are used to rank the formalizations. We compare the results of the LLM judges with those of the human evaluators and report the results in Table~\ref{tab:pair}.

\textbf{Fine-grained LLM judges perform consistently better than coarse-grained judges.} In Table~\ref{tab:pair}, across all LLMs, the method of using LLM judges with OAPs achieves the highest exact match score in 12 cases and the highest pairwise match score in 17 out of 20 cases compared to the other two settings. On miniF2F, all settings with the highest individual scores are achieved via OAPs. On ProofNet, three out of four highest individual scores are also achieved via OAPs. Together, these results demonstrate the superiority of the OAP setting and support our claim that using fine-grained aspects in LLM-based judges for autoformalization evaluation is more accurate than coarse-grained approaches.

\textbf{Fine-grained LLM judges can reach the same level of human agreement on autoformalization evaluation.} Comparing results in Table~\ref{tab:pair} with human agreement scores, with OAP settings, LLM judges can reach performance levels that are comparable to human inter-annotator agreement, particularly in pairwise matching where their scores approach or exceed the agreement between human annotators on both miniF2F and ProofNet. This indicates that decomposing evaluation into fine-grained aspects enables LLM judges to more closely approximate human judgment behavior, improving consistency and reliability in autoformalization assessment.

\begin{table*}[t]
  \centering
  \begin{tabular}{llcccccc}
    \toprule
    && \multicolumn{2}{c}{Isabelle (miniF2F)} & \multicolumn{2}{c}{Lean4 (miniF2F)} & \multicolumn{2}{c}{Lean4 (ProofNet)}\\
    Aspect & Judge & GT & ZS & GT & ZS & GT & ZS\\
    \midrule
    LP & Human Expert & $0.24_{\pm0.43}$ & $0.80_{\pm0.40}$ & $0.40_{\pm0.49}$ & $0.80_{\pm0.40}$ & $0.68_{\pm0.47}$ & $0.40_{\pm0.49}$\\
    MC & Human Expert & $0.34_{\pm0.47}$ & $0.74_{\pm0.44}$ & $0.40_{\pm0.49}$ & $0.52_{\pm0.50}$& $0.68_{\pm0.48}$ & $0.40_{\pm0.49}$\\
    FQ & Human Expert & $0.76_{\pm0.43}$ & $0.46_{\pm0.50}$ & $0.92_{\pm0.27}$& $0.56_{\pm0.50}$& $0.92_{\pm0.27}$ & $0.60_{\pm0.49}$\\
    FV & Theorem Prover & $1.00_{\pm0.00}$ & $0.52_{\pm0.50}$& $1.00_{\pm0.00}$& $0.20_{\pm0.40}$& $1.00_{\pm0.00}$ & $0.12_{\pm0.32}$\\
    OA & Human Expert & $4.06_{\pm0.58}$ & $4.00_{\pm0.77}$ & $4.04_{\pm0.53}$& $3.64_{\pm0.93}$& $4.48_{\pm0.64}$ & $3.56_{\pm0.57}$\\
    \bottomrule \\
  \end{tabular}
  \caption{Human evaluation of formalizations on subsets of validation sets. Results are reported as mean$\pm$ standard deviation across samples. (\textbf{GT}): Ground-Truth formalization; (\textbf{ZS}): GPT-4.1 Zero-Shot formalization; (\textbf{LP}): Logical Preservation; (\textbf{MC}): Mathematical Consistency; (\textbf{FQ}): Formal Quality; (\textbf{FV}): Formal Validity; (\textbf{OA}): Overall Assessment.}
  \label{tab:human}
\end{table*}

\textbf{Smaller fine-grained LLM judges can exceed the performance of coarse-grained larger LLM judges.} Across both miniF2F and ProofNet, we observe that GPT-4.1-mini equipped with fine-grained OAP-based evaluation consistently outperforms larger models using coarse-grained evaluation setups. The decomposition of evaluation into interpretable aspects enables smaller judges to better capture localized correctness signals, reducing the ambiguity inherent in single-pass judgments. These results suggest that evaluation quality is driven not only by model scale, but also by the structure of the judging process, with fine-grained decomposition providing a more effective approach for LLM-based assessment.

\textbf{Computational cost can be reduced by selecting smaller effective LLMs while maintaining strong performance.} On miniF2F, under the OAP setting, GPT-4.1-mini even outperforms its larger counterpart GPT-4.1 and a more recent model, GPT-5.4. On ProofNet, while GPT-4.1-mini is not the top-performing model, it still achieves competitive performance. Since the OAP setting requires multiple LLM judge passes, these results suggest a favorable trade-off between computational cost and performance when using smaller LLMs with OAPs for autoformalization evaluation.

\begin{figure*}[t]
    \centering
    \begin{subfigure}{0.24\textwidth} 
      \centering
      \includegraphics[width=\textwidth]{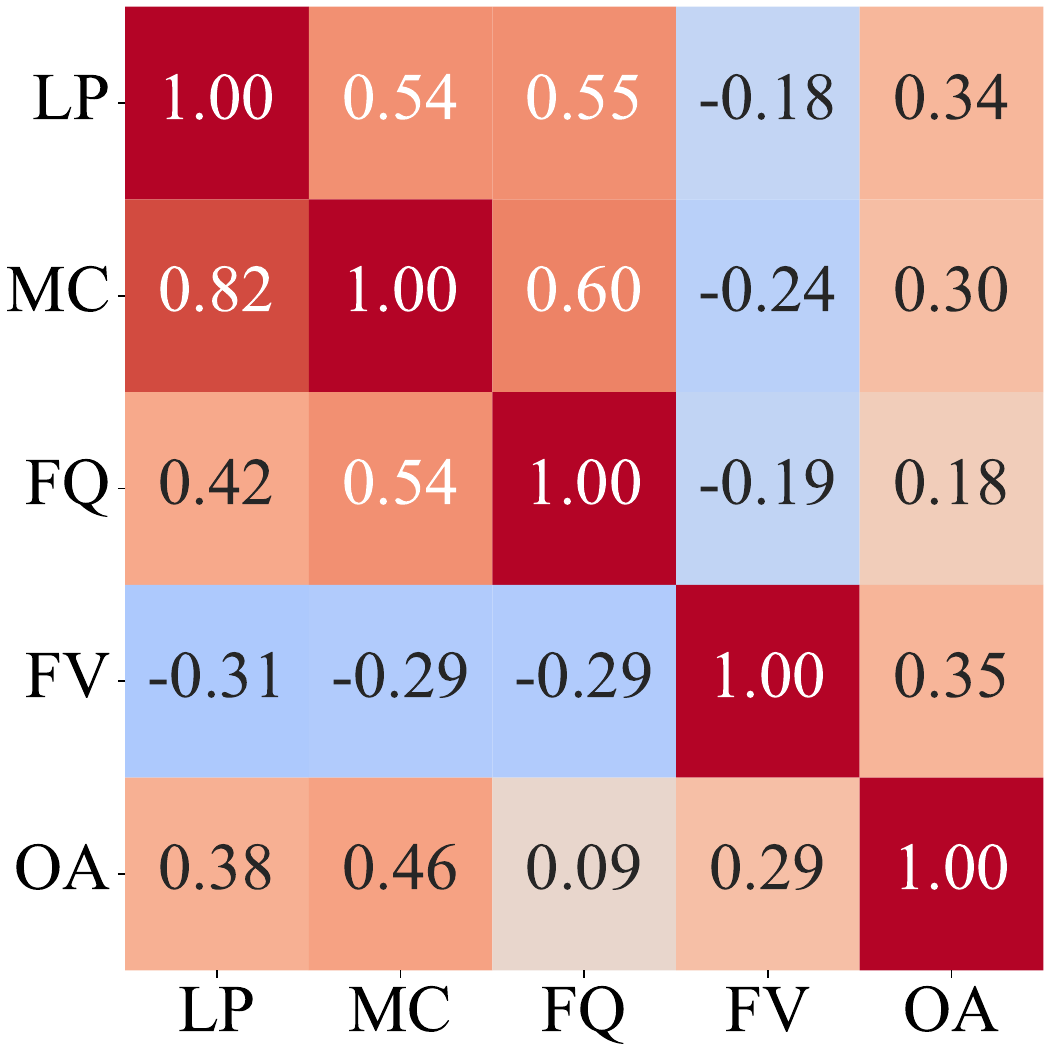}
      \caption{Qwen2.5-Coder-7B}
      \label{fig:qwen}
    \end{subfigure}
    \begin{subfigure}{0.24\textwidth} 
      \centering
      \includegraphics[width=\textwidth]{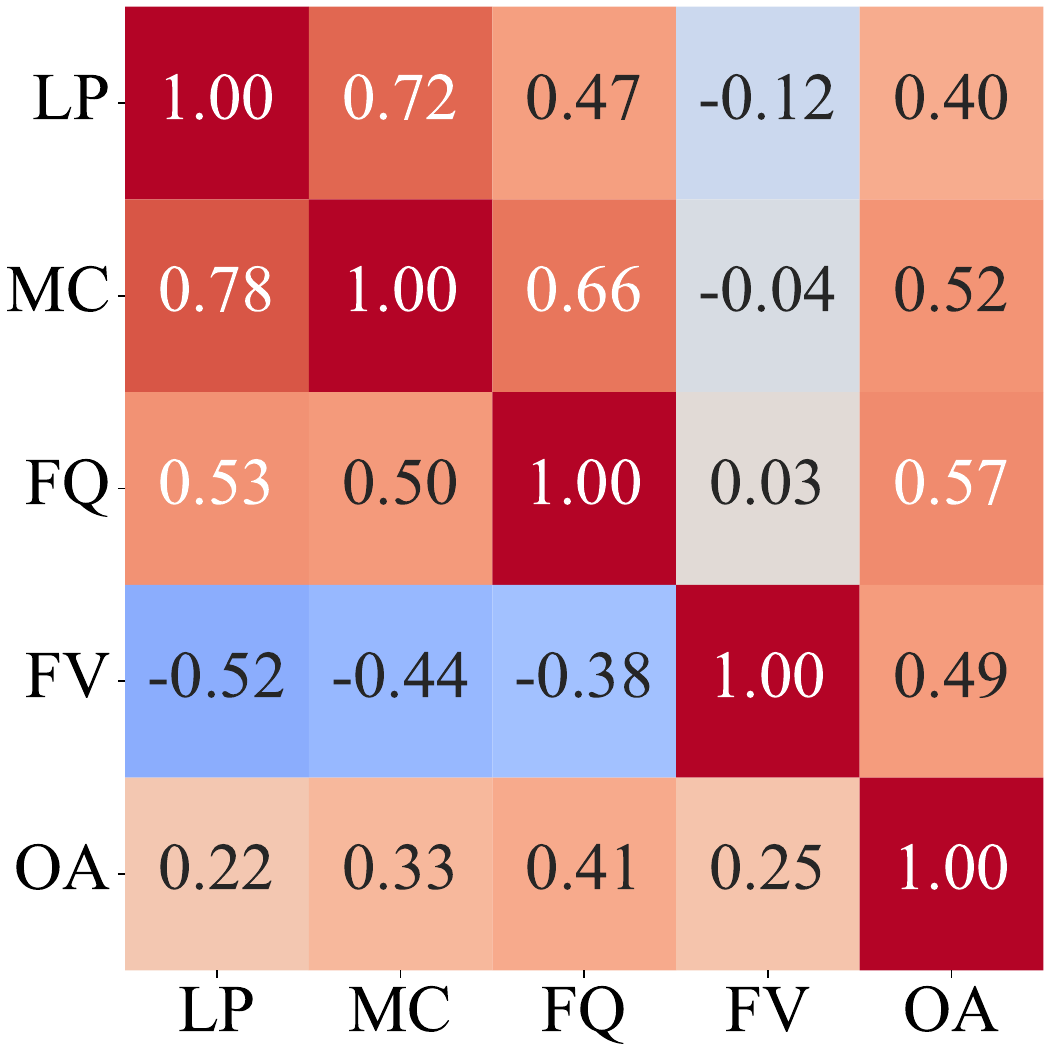}
      \caption{GPT-4.1-mini}
      \label{fig:gpt-mini}
    \end{subfigure}
    \begin{subfigure}{0.24\textwidth} 
      \centering
      \includegraphics[width=\textwidth]{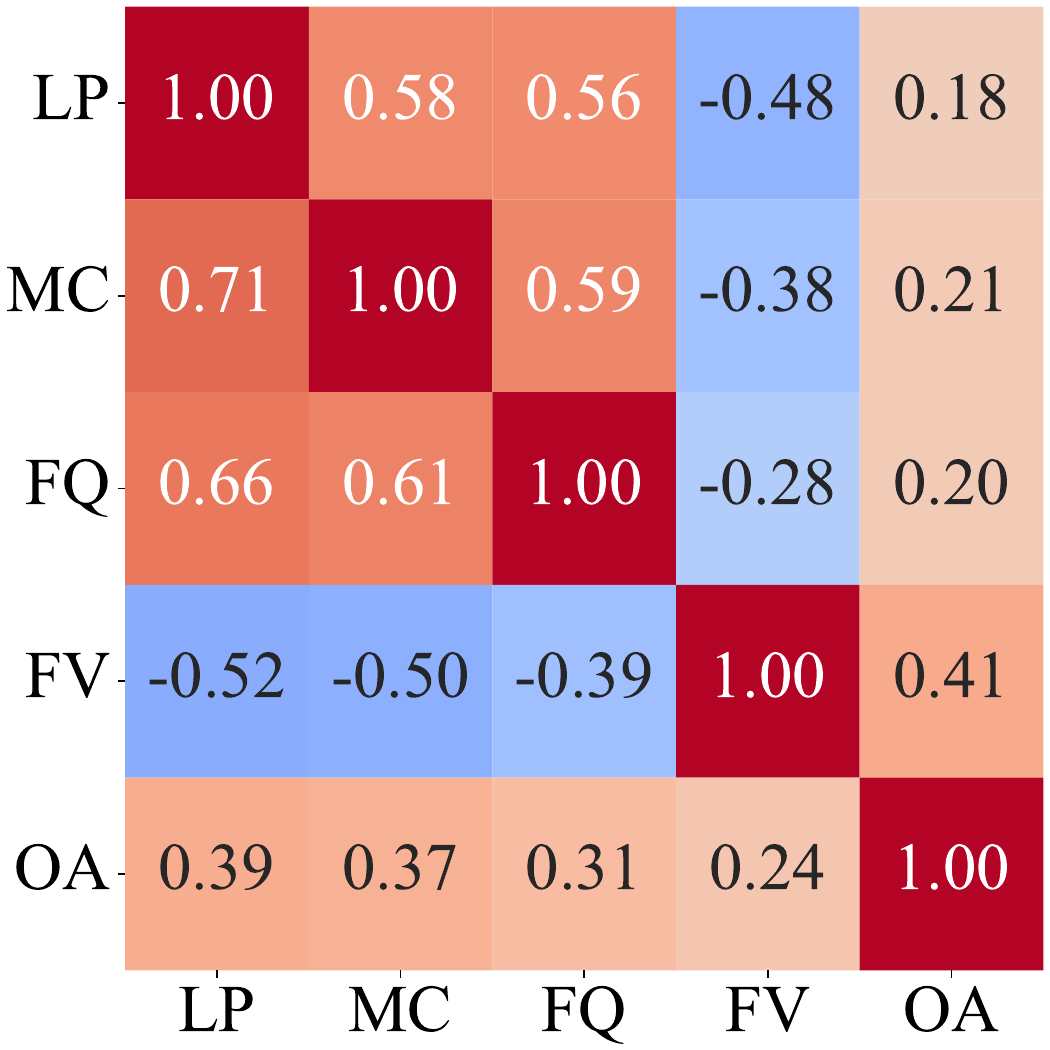}
      \caption{GPT-4.1}
      \label{fig:gpt}
    \end{subfigure}
    \begin{subfigure}{0.24\textwidth} 
      \centering
      \includegraphics[width=\textwidth]{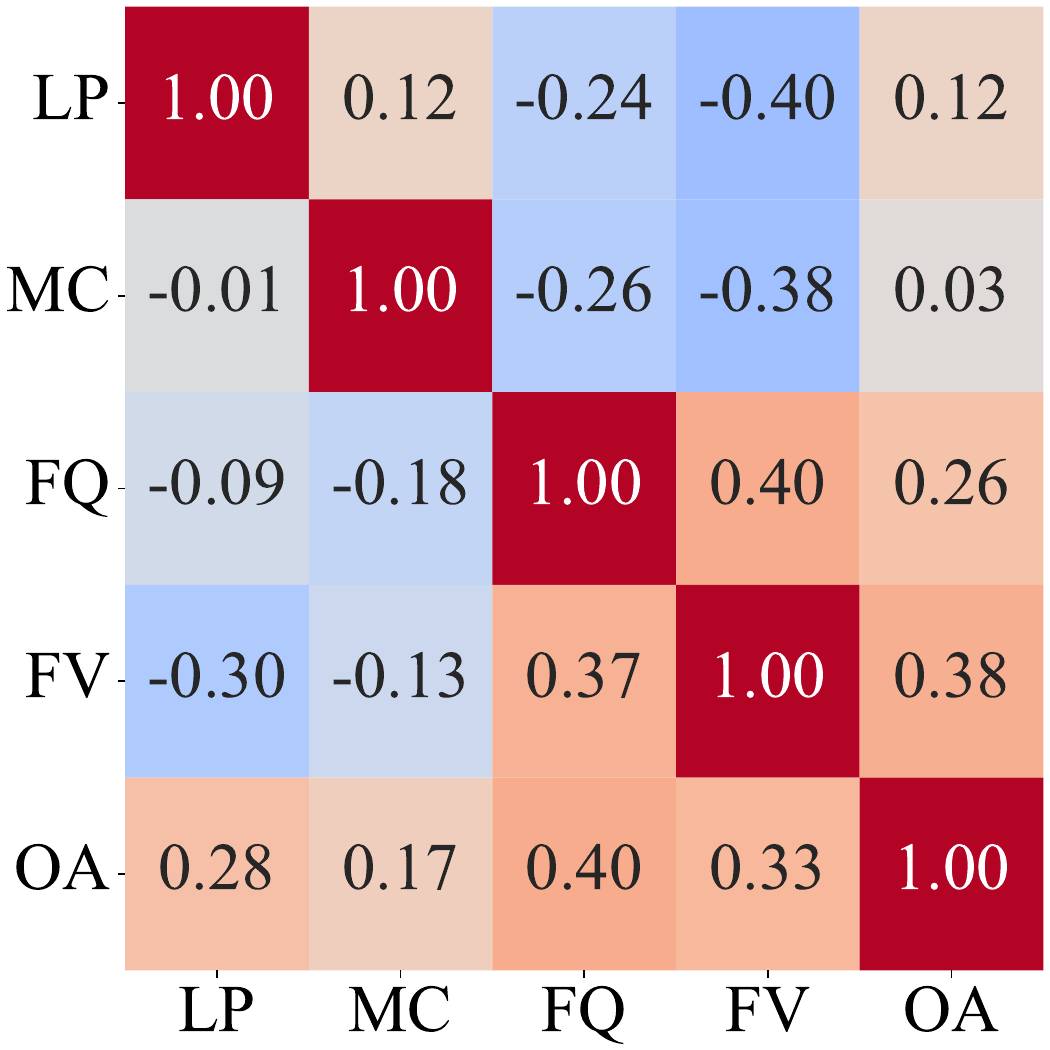}
      \caption{Human}
      \label{fig:human}
    \end{subfigure}
    \caption{Pearson correlation coefficients between evaluation aspects and overall assessments across judges for flawed formalizations. LLM judges and human evaluators assess the Tier-2 aspects logical preservation (LP), mathematical consistency (MC), and formal quality (FQ), while theorem provers assess formal validity (FV). The lower triangular matrix shows results for Lean4, and the upper triangular matrix shows results for Isabelle/HOL.}
    \label{fig:corr}
\end{figure*}

\subsection{Contribution of Evaluation Aspects in LLM and Human Judgments}
For a perfect formalization, all evaluation aspects should receive positive assessments. For formalizations judged to be flawed, however, it is important to understand how LLM judges and human evaluators differ in how individual evaluation aspects contribute to the final assessment. To enable a more detailed comparison between human evaluators and LLM judges, we focused on natural language statements that were likely to result in formalization errors, even in human-written formalizations. To this end, we first used GPT-4.1 to estimate whether the human-written ground-truth formalizations in the validation datasets correctly represented their corresponding natural-language statements. The estimated correctness rates were 76.64\% for miniF2F Isabelle/HOL, 82.79\% for miniF2F Lean4, and 64.84\% for ProofNet Lean4. Based on these estimates, we randomly sampled 50, 25, and 25 natural-language statements from miniF2F Isabelle/HOL, miniF2F Lean4, and ProofNet Lean4, respectively, whose ground-truth formalizations were identified as potentially incorrect.

For each selected statement, both the ground-truth formalization and the GPT-4.1 zero-shot formalization are evaluated by one human expert and LLM judges. The human expert and LLM judges evaluated the formalizations with respect to LP, MC, and FQ using binary judgments, together with an overall assessment on a 5-point scale (criteria in Table~\ref{tab:anno} in Appendix~\ref{app:prompt}). FV, as verified by the corresponding theorem prover, was also taken into account during the human evaluation and the overall assessment from LLM judges.

We acknowledge the potential subjectivity inherent in human evaluation, particularly regarding the relative importance assigned to different evaluation aspects. Although a multi-annotator study could provide additional insights by analyzing aspect contributions separately for each annotator, conducting such an analysis would substantially increase the complexity of the evaluation procedure and is beyond the scope of this work. We therefore employ a single expert evaluator to provide a consistent and practically reliable reference for comparison. We report the mean and standard deviation of the human assessments in Table~\ref{tab:human}. As expected, the annotated formalizations are not perfect, exhibiting errors across all Tier-2 evaluation aspects.

\subsubsection{Results \& Discussions}
We calculate Pearson correlation coefficients between Tier-2 aspect evaluations and overall assessments from LLM judges and the human evaluator on flawed formalizations identified by the human evaluator. Results for Lean4 and Isabelle/HOL are reported in Figure~\ref{fig:corr}.

\textbf{Human and LLM judges exhibit different correlation patterns among Tier-2 aspects when evaluating flawed formalizations.} As shown in Figures~\ref{fig:qwen}, \ref{fig:gpt-mini}, and \ref{fig:gpt}, LLM judges generally exhibit moderate positive correlations among the soft aspects, logical preservation (LP), mathematical consistency (MC), and formal quality (FQ), whereas human evaluators show weak or slightly negative correlations between these aspects. This suggests that LLM judges may have difficulty distinguishing subtle differences between aspects as humans do, leading to more entangled judgments. In contrast, the hard aspect of formal validity (FV) shows weak or negative correlations with LP and MC for both human and LLM judges. This indicates that semantic alignment issues are largely distinct from syntactic correctness: a formally valid formalization can still be misaligned with the source mathematics. Both humans and LLMs appear capable of recognizing this distinction. Human evaluators also exhibit a positive correlation between FQ and FV, reflecting the tendency for well-structured formalizations to be both more readable and more likely to satisfy syntactic requirements. This relationship is largely absent for LLM judges, suggesting that humans and LLMs rely on different notions of formal quality when assessing flawed formalizations.

\textbf{All Tier-2 aspects contribute to the overall assessment for both human and LLM judges.} All aspects exhibit positive correlations with the overall assessment, indicating that these dimensions serve as informative signals for formalization quality. However, the strength of these relationships, as reflected in the correlation coefficients, varies across evaluators and formal languages. For Lean4, both Qwen2.5-Coder-7B and GPT-4.1 assign relatively higher associations to LP and MC compared to FQ and FV, whereas GPT-4.1-mini and human evaluators show the strongest association with FQ. For Isabelle/HOL, Qwen2.5-Coder-7B, GPT-4.1, and human evaluators place the strongest association on FV, while GPT-4.1-mini exhibits a more uniform pattern across all aspects. These findings indicate that evaluators do not rely on a single dominant aspect in a uniform manner but instead integrate multiple factors with varying degrees of emphasis depending on the evaluator type and formal language.

\begin{table*}[t]
  \centering
  \begin{tabular}{llccccc}
    \toprule
    Prompting & Model & LP & MC & FV & FQ & OA\\
    \midrule
    \multicolumn{4}{l}{\textit{miniF2F-Test Isabelle/HOL}}\\
    \midrule
    Zero-Shot & GPT-4.1 & 0.939 & 0.963 & 0.643 & \underline{\textbf{0.953}} & 0.874\\
    8-Shot & GPT-4.1 & \underline{\textbf{0.945}} & \underline{\textbf{0.986}} & \underline{\textbf{0.746}} & 0.949 & \underline{\textbf{0.906}}\\
    \mycdashline{1-7}
    8-Shot & Mistral-7B & 0.418 & 0.624 & 0.201 & 0.480 & 0.431\\
    8-Shot & Deepseek-Math-7B & 0.492 & 0.698 & 0.324 & 0.512 & 0.506\\
    8-Shot & Qwen2.5-7B & \underline{0.651} & \underline{0.831} & \underline{0.590} & \underline{0.740} & \underline{0.703}\\
    \midrule
    \multicolumn{4}{l}{\textit{miniF2F-Test Lean4}}\\
    \midrule
    Zero-Shot & GPT-4.1 & \underline{\textbf{0.943}} & \underline{\textbf{0.986}} & 0.463 & 0.955 & 0.837\\
    8-Shot & GPT-4.1 & 0.924 & 0.985 & \underline{\textbf{0.697}} & \underline{\textbf{0.957}} & \underline{\textbf{0.891}}\\
    \mycdashline{1-7}
    8-Shot & Mistral-7B & 0.434 & 0.634 & 0.430 & 0.492 & 0.498\\
    8-Shot & Deepseek-Math-7B & 0.522 & 0.750 & 0.561 & 0.641 & 0.619\\
    8-Shot & Qwen2.5-7B & \underline{0.627} & \underline{0.842} & \underline{0.586} & \underline{0.758} & \underline{0.703}\\
    \midrule
    \multicolumn{4}{l}{\textit{ProofNet-Test Lean4}}\\
    \midrule
    Zero-Shot & GPT-4.1 & 0.923 & 0.980 & 0.137 & 0.945 & 0.746\\
    8-Shot & GPT-4.1 & \underline{\textbf{0.963}} & \underline{\textbf{0.993}} & \underline{\textbf{0.423}} & \underline{\textbf{0.967}} & \underline{\textbf{0.836}}\\
    \mycdashline{1-7}
    8-Shot & Mistral-7B & 0.317 & 0.604 & 0.082 & 0.393 & 0.349\\
    8-Shot & Deepseek-Math-7B & \underline{0.569} & \underline{0.793} & \underline{0.192} & 0.585 & \underline{0.535}\\
    8-Shot & Qwen2.5-7B & 0.559 & 0.769 & 0.099 & \underline{0.613} & 0.510\\
    \bottomrule
  \end{tabular}
  \caption{Autoformalization evaluation with GPT-4.1-mini in the OAP setting. (\textbf{LP}): Logical Preservation; (\textbf{MC}): Mathematical Consistency; (\textbf{FQ}): Formal Quality; (\textbf{FV}): Formal Validity; (\textbf{OA}): Overall Assessment.}
  \label{tab:auto_test}
\end{table*}

\subsection{Distinguishing Performance}
We use the GPT-4.1-mini OAP method to assess the autoformalization performance of models on datasets. We evaluate the zero-shot setting with GPT-4.1, and the few-shot (8 exemplars) setting with GPT-4.1, Mistral-7B~\citep{jiang2023mistral}, Deepseek-Math-7B~\citep{shao2024deepseekmath}, and Qwen2.5-7B~\citep{qwen2025qwen25technicalreport}. We report scores in Table~\ref{tab:auto_test}. 

GPT-4.1 in the few-shot setting generally demonstrates better performance compared to the zero-shot setting. However, \textbf{this performance gain is primarily driven by improvements in formal validity, indicating that few-shot exemplars mainly help the model produce syntactically valid formalizations.} These findings suggest that our proposed evaluation method offers more informative insights into specific areas of improvement.

Among the 7B models, \textbf{Qwen2.5-7B achieves the best performance across all metrics on miniF2F}, while on ProofNet, the results are more varied, with Deepseek-Math-7B achieving the highest overall assessment (OA). Additionally, the performance of 7B LLMs on miniF2F and ProofNet in Lean4 still shows a significant gap in OA, indicating that formalizing complex statements remains challenging for these models.



\subsubsection{Biases Toward Self-Generated Outputs}
We investigate the use of different LLMs as OAP judges for autoformalization evaluation. Specifically, we use GPT-4.1-mini and Qwen2.5-Coder-7B to evaluate the few-shot autoformalizations produced by GPT-4.1 and Qwen2.5-7B on the miniF2F test set. Scores for Tier-2 aspects and overall assessments are reported in Table~\ref{tab:judge_diff}.

\begin{table*}[t]
    \centering
    \begin{tabular}{llcccccccc}
    \toprule
    Autoformalizer & Judge & LP & MC & FQ & OA\\
    \midrule
    \multicolumn{4}{l}{\textit{Lean4}}\\
    \midrule
    GPT-4.1 & GPT-4.1-Mini & \underline{0.924} & \underline{0.985} & \underline{0.957} & \underline{0.891}\\
    Qwen2.5-7B & GPT-4.1-Mini & 0.627 & 0.842 & 0.758 & 0.703\\
    \mycdashline{1-6}
    GPT-4.1 & Qwen2.5-Coder-7B & \underline{0.703} & \underline{0.757} & \underline{0.328} & \underline{0.621}\\    
    Qwen2.5-7B & Qwen2.5-Coder-7B & 0.473 & 0.616 & 0.170 & 0.461\\
    \midrule
    \multicolumn{4}{l}{\textit{Isabelle/HOL}}\\
    \midrule
    GPT-4.1 & GPT-4.1-Mini & \underline{0.945} & \underline{0.986} & \underline{0.949} & \underline{0.906}\\
    Qwen2.5-7B & GPT-4.1-Mini & 0.651 & 0.831 & 0.740 & 0.703\\
    \mycdashline{1-6}
    GPT-4.1 & Qwen2.5-Coder-7B & \underline{0.847} & \underline{0.833} & \underline{0.666} & \underline{0.773}\\
    Qwen2.5-7B & Qwen2.5-Coder-7B & 0.666 & 0.717 & 0.518 & 0.623\\
    \bottomrule
    \end{tabular}
  \caption{Using different judges for evaluation on the miniF2F test set. LLMs are used in few-shot setting.}
  \label{tab:judge_diff}
\end{table*}

Qwen2.5-Coder-7B generally assigns lower scores compared to GPT-4.1-mini, suggesting that absolute score values may vary across judge models. Nevertheless, \textbf{Qwen2.5-Coder-7B consistently rates GPT-4.1's autoformalizations higher than those of Qwen2.5-7B across both formal languages, indicating a degree of robustness in the indicative power of our evaluation approach}. The score differences between the two judges are more pronounced on Lean4 than on Isabelle/HOL, which may reflect biases in the training data of Qwen2.5-Coder-7B. \textbf{At present, we find no evidence that the LLM-as-a-judge framework favors autoformalizations generated by the same model series.}

\subsubsection{Randomness in LLM Judges}\label{sec:random}
\begin{figure}
  \centering
  \includegraphics[width=0.46\textwidth]{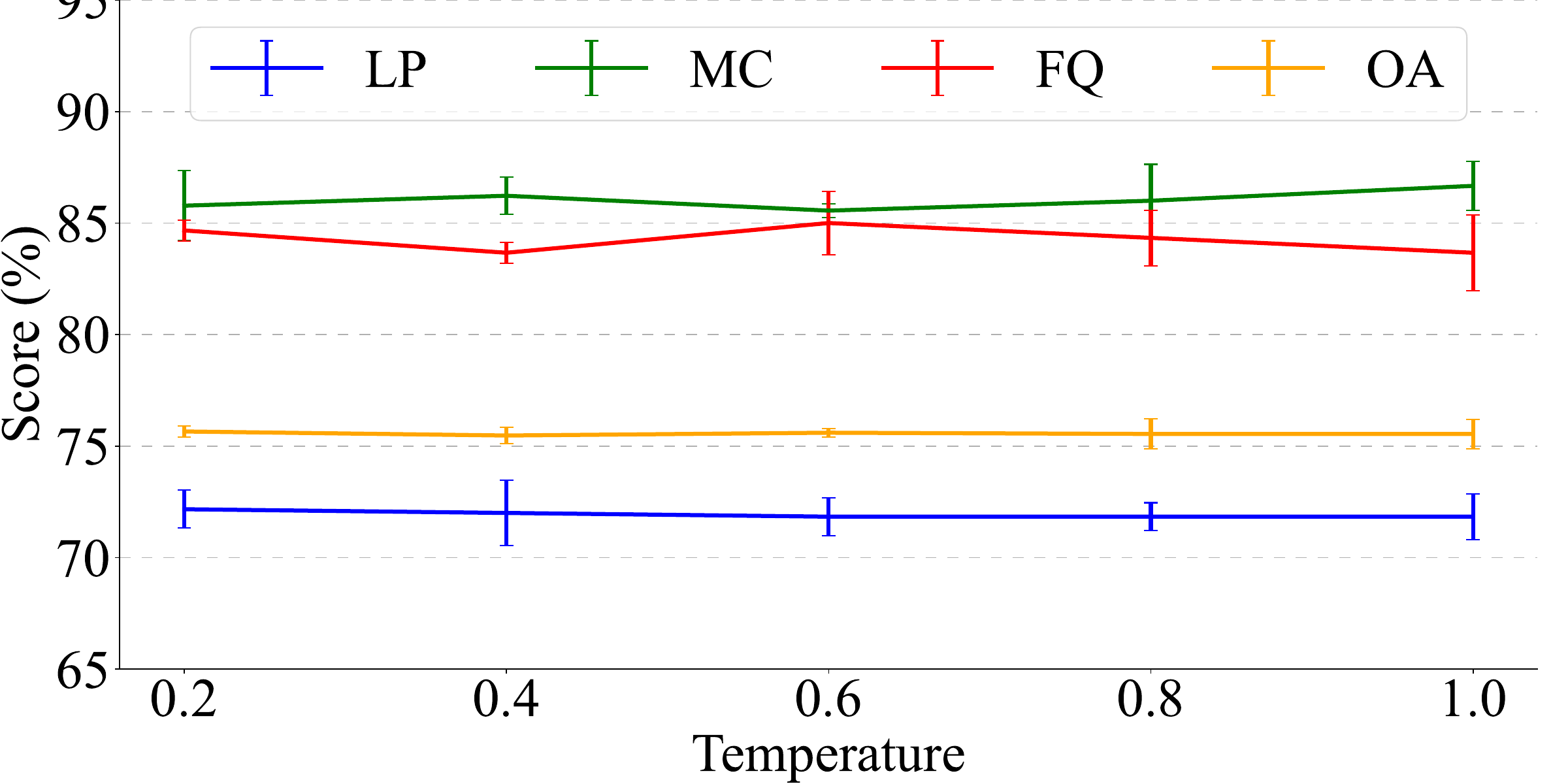}
  \caption{Error bars of GPT-4.1-mini with the OAP setting running with different temperatures.}
  \label{fig:robust}
\end{figure}

We explore how the randomness introduced during the generation of LLMs influences the scores with GPT-4.1-mini OAP-EW. We randomly select 50 Isabelle/HOL formalizations from the miniF2F test set using Qwen2.5-7B with 8-shot prompting. We perform 3 runs for 5 different temperatures (0.2, 0.4, 0.6, 0.8, 1.0) and report the mean and variance of the scores in Figure~\ref{fig:robust}. In general, the mean of the MC and FQ scores fluctuates as the temperature increases, whereas the LP and OA scores remain relatively stable. \textbf{While LP, MC, and FQ scores exhibit some variance across multiple runs, the OA scores show consistent stability.}

\section{Related Work}

\subsection{Autoformalization}
Autoformalization bridges the gap between natural and formal languages, enabling natural language inputs to be translated into formal representations. This task has seen promising developments across various domains, such as natural language inference~\citep{quan-etal-2024-enhancing,quan-etal-2024-verification} and automated theorem proving~\citep{jiang2023draft, tarrach2024more}. LLMs have shown efficacy in Isabelle~\citep{wu2022autoformalization}, and retrieval-augmented generation has further enhanced performance in Isabelle~\citep{zhang-etal-2024-consistent} and Lean~\citep{yang2023leandojotheoremprovingretrievalaugmented,liu2025rethinking}. In parallel, \citet{jiang2024multilanguage} and \citet{liu2025atlasautoformalizingtheoremslifting} developed data generation pipelines for constructing large-scale parallel corpora of theorem statements. On the evaluation front, \citet{poiroux2025improvingautoformalizationusingtype} proposed scalable type-check filtering with self-consistency, while \citet{lu2024formalalignautomatedalignmentevaluation} trained a model to assess semantic alignment between informal and formal statements. Despite these advances, standardized and comprehensive evaluation metrics for autoformalization remain under-explored. Our work seeks to address this critical gap.

\subsection{LLM-as-a-Judge}
LLM-as-a-Judge~\citep{zheng2023judging} was proposed for evaluating multi-turn questions and dialogues and has become an increasingly popular approach for assessing NLP tasks using LLMs. In the domain of mathematical evaluation, \citet{zhang2025deeptheoremadvancingllmreasoning} explored the use of LLM judges for scoring natural language proofs across four dimensions. However, limited research has been conducted on employing LLM-as-a-Judge in the field of formal mathematics. Our work takes a step in this direction.




\section{Conclusion}
In this work, we address the under-explored challenge of evaluating autoformalization by introducing a fine-grained taxonomy that captures multiple aspects of a candidate formalization and propose a linear evaluation model to synthesize core properties into a single overall assessment score. 

Through human evaluation and empirical analysis, we demonstrate that this automatic evaluation model provides meaningful insights into formalization performance. Furthermore, we show the feasibility of an EFG ensemble of LLM judges, both directly and via OAPs. Our experiments reveal that fine-grained LLM-based judgments align more closely with human assessments than coarse-grained judgements in terms of ranking. Lightweight LLMs like GPT-4.1-mini, when guided by OAPs, can serve as effective and cost-efficient judges. 

Our findings contribute new insights and tools for automatic and fine-grained evaluation of autoformalization systems, paving the way for more systematic benchmarking and future improvements in this emerging area. Further directions could focus on fine-tuning smaller open-source LLMs specialized as judges for autoformalization evaluation and using LLM explanations of judgments to improve autoformalization performance.

\section*{Limitations}
As with other studies involving human annotations, some of our results may be affected by subjective biases. However, we attempt to limit subjectivity by leveraging the annotators with background knowledge in both AI and formal languages, capable of articulating the rationale behind the annotators' decisions. Additionally, the scale of human-annotated samples is relatively small due to the complexity involved in the annotation. While expanding the annotation would likely yield more robust and generalizable results, we do not expect it to alter the main findings. Moreover, the set of operable atomic properties and LLMs used in our approach is not exhaustive. Although alternative configurations might yield slightly different results, we believe they would not substantially impact the core conclusions of the study.

\section*{Acknowledgments}

Thanks to Lulu Du for her valuable feedback.


\bibliography{custom}

\appendix

\section{Human Annotation Criteria and LLM Prompts}\label{app:prompt}
We provide the detailed human annotation criteria in Table~\ref{tab:anno}. A binary scheme is used for logical preservation, mathematical consistency, and formal quality, while overall assessment is rated on a 5-point scale. Since mathematics is a rigorous discipline and formal languages are designed to reflect this rigor, any human with sufficient knowledge of mathematics and formal language should be able to apply these criteria consistently and reach similar judgments. We provide the prompts in Table~\ref{tab:prompt}.
\begin{table*}[t]
  \centering
  \small
  \begin{tabular}{p{0.95\textwidth}}
    \toprule
    \textbf{Logical Preservation (LP)}\newline
    From your point of view, does the formal code capture the logical structure and content of the natural language statement?\newline
    \textbf{1}: Yes\quad\textbf{0}: No\\
    \midrule
    \textbf{Mathematical Consistency (MC)}\newline
    From your point of view, does the formal code accurately represent mathematical objects and operations present in the natural language statement?\newline
    \textbf{1}: Yes\quad\textbf{0}: No\\
    \midrule
    \textbf{Formal Quality (FQ)}\newline
    From your point of view, does the formalized code alone demonstrate high quality in terms of structural clarity, readability, and usefulness?\newline
    \textbf{1}: Yes\quad\textbf{0}: No\\
    \midrule
    \textbf{Overall Assessment (OA)}\newline
    What would you rate this formal code as the formalization of the natural language statement?\newline
    \textbf{5}: Perfect formalization – fully accurate and complete with no issues.\newline
    \textbf{4}: Good formalization – minor flaws are present, but they are easy to fix.\newline
    \textbf{3}: Acceptable formalization – captures the intent but may require some time and effort to correct.\newline
    \textbf{2}: Flawed formalization – contains major issues that require significant time and effort to correct.\newline
    \textbf{1}: Poor formalization – completely fails to reflect the intended meaning of the natural language statement.\\
    \bottomrule\\
  \end{tabular}
  \caption{Annotation criteria}
  \label{tab:anno}
\end{table*}

\section{Operable Atomic Properties}\label{app:oap}
With the notation of $s$ as the original natural language statement and $\phi$ as a candidate formalization, we define operable atomic properties as follows:

\begin{enumerate}
	\item \textbf{Pre-arg Structure:} $\phi$ reflects the inherent predicate-argument structure of the statement $s$.
	\item \textbf{Quantification:} $\phi$ accurately formalizes all quantifiers, such as universal and existential, present in $s$.
	\item \textbf{Formula:} All mathematical formulas and expressions in $s$, such as equations and inequalities, are correctly and completely represented in $\phi$.
	\item \textbf{Relation:} The logical and mathematical relationships between propositions in \(s\) are preserved in $\phi$.
	\item \textbf{Concept:} All mathematical concepts mentioned in $s$, such as integers, fractions, real or complex numbers, derivatives, integrals, vectors, matrices, probabilities, expectations, and variances, are correctly formalized in $\phi$.
	\item \textbf{Constant:} All mathematical constants mentioned in $s$, such as 1, $\frac{2}{3}$, $\pi$, $e$, are properly included in $\phi$.
	\item \textbf{Operator:} All mathematical operators used in $s$, such as addition, subtraction, multiplication, division, summation, exponentiation, and product, are correctly represented in $\phi$.
	\item \textbf{Conciseness:} $\phi$ is expressed in a minimal, non-redundant form, avoiding unnecessary repetition or complexity.
	\item \textbf{Logical Consistency:} $\phi$ is internally coherent and contains no contradictions under the logical rules of the formal system.
\end{enumerate}

\section{Examples of Incorrect LLM Judgments}\label{app:example}
We conduct a qualitative study on cases where GPT-4.1 misjudges logical preservation and mathematical consistency in formalizations. We provide two examples of misjudged logical preservation and two examples of misjudged mathematical consistency by GPT-4.1 in Table~\ref{tab:case_1} and Table~\ref{tab:case_2}, and in Table~\ref{tab:case_3} and Table~\ref{tab:case_4}, respectively. GPT-4.1 often provides reasonable and correct explanations of the formalizations, and its judgments generally align with those explanations. The primary reason for incorrect judgments is the overlap between the logical and mathematical aspects of formalizations, which makes it difficult for GPT-4.1 to clearly distinguish the subtle differences between them through prompt-based interaction.

\begin{table*}[t]
  \centering
  \small
  \begin{tabular}{p{0.14\textwidth} p{0.8\textwidth}}
    \toprule
    NL Statement & 
    Let \[f(x) =
    \begin{cases}
    x^2+9 &\mathrm{if }x<-5, \\
    3x-8&\mathrm{if }x\ge-5.
    \end{cases}\]
    If $f(x)=10$, find the sum of all possible values of $x$. Show that it is 6.\\
    \mycdashline{1-2}
    Human\newline Formalization &
    \begin{lstlisting}[language=isabelle,aboveskip=-7pt,belowskip=-12pt]
theorem mathd_algebra_149:
  fixes f :: "real \<Rightarrow> real"
  assumes "\<forall> x < -5. f x = x^2 + 5"
    and "\<forall> x \<ge> -5. f x = 3 * x -8"
  shows "(\<Sum> k \<in> (f -` {10}). k) = 6" 
  sorry
    \end{lstlisting}\\
    \mycdashline{1-2}
    Aspect & Logical Preservation\\
    \mycdashline{1-2}
    Human & Judgment: True\newline
    Explanation: The issue of having `+5' instead of `+9' in the formalization does not indicate that the formalization fails to capture the logic of the natural language statement; rather, it suggests a misrepresentation of a mathematical constant.\\
    \mycdashline{1-2}
    GPT-4.1 &
    Explanation: The formal code does not fully capture the logical structure and content of the natural language statement. The natural language statement defines a piecewise function  \( f(x)  \) with  \( f(x) = x^2 + 9  \) for  \( x < -5  \) and  \( f(x) = 3x - 8  \) for  \( x  \geq -5  \). However, in the formal code, the first case is written as  \( f x = x^2 + 5  \) instead of  \( x^2 + 9  \), which is a critical error. This changes the function being analyzed and thus the set of solutions to  \( f(x) = 10  \). The rest of the structure, including the use of the preimage and summing over the solutions, is correct, but the incorrect function definition means the code does not faithfully represent the original problem.\newline Judgment: False\\
    \bottomrule\\
  \end{tabular}
  \caption{Example 1 of incorrect LLM judgment in Isabelle/HOL.}
  \label{tab:case_1}
\end{table*}

\begin{table*}[t]
  \centering
  \small
  \begin{tabular}{p{0.14\textwidth} p{0.8\textwidth}}
    \toprule
    NL Statement & How many distinct, non-equilateral triangles with a perimeter of 60 units have integer side lengths $a$, $b$, and $c$ such that $a$, $b$, $c$ is an arithmetic sequence? Show that it is 9.\\
    \mycdashline{1-2}
    GPT-4.1\newline Formalization &
    \begin{lstlisting}[language=isabelle,aboveskip=-7pt,belowskip=-12pt]
lemma count_non_equilateral_triangles_arith_seq:
  "card { (a, b, c) :: nat × nat × nat.
      a < b ∧ b < c ∧
      a + b + c = 60 ∧
      (∃d::nat. d > 0 ∧ b = a + d ∧ c = a + 2*d) ∧
      ¬ (a = b ∨ b = c ∨ a = c) ∧
      a + b > c ∧ a + c > b ∧ b + c > a
    } = 9"
  sorry
    \end{lstlisting}\\
    \mycdashline{1-2}
    Aspect & Logical Preservation\\
    \mycdashline{1-2}
    Human & Judgment: False\newline
    Explanation: Non-equilateral triangle should be $\neg(a=b\wedge b = c\wedge a = c)$ rather than $\neg(a=b\vee b = c\vee a = c)$ which represent non-isosceles triangle.\\
    \mycdashline{1-2}
    GPT-4.1 &
    Explanation: The formalization captures most of the logical structure and content of the natural language statement. It defines the set of triples \((a, b, c)\) of natural numbers such that \(a < b < c\), the perimeter is 60, the sides form an arithmetic sequence (with \(b = a + d\), \(c = a + 2d\), \(d > 0\)), the triangle is non-equilateral (\(\neg (a = b \vee b = c \vee a = c)\)), and the triangle inequalities are satisfied. The cardinality of this set is asserted to be 9, matching the "show that it is 9" part. The only minor issue is that the triangle inequalities are redundant given \(a < b < c\) and the arithmetic sequence, but their inclusion does not harm correctness. The formalization does not miss any logical aspect of the natural language statement.\newline
    Judgement: True\\
    \bottomrule\\
  \end{tabular}
  \caption{Example 2 of incorrect LLM judgment in Isabelle/HOL.}
  \label{tab:case_2}
\end{table*}

\begin{table*}[t]
  \centering
  \small
  \begin{tabular}{p{0.14\textwidth} p{0.8\textwidth}}
    \toprule
    NL Statement & Let $H \leq K \leq G$. Prove that $|G: H|=|G: K| \cdot|K: H|$ (do not assume $G$ is finite).\\
    \mycdashline{1-2}
    GPT-4.1\newline Formalization &
    \begin{lstlisting}[language=isabelle,aboveskip=-7pt,belowskip=-12pt]
import Mathlib.GroupTheory.Index

variable {G : Type*} [Group G] (H K : Subgroup G) (hHK : H ≤ K)

theorem index_mul_index_eq_index (hHK : H ≤ K) : 
  Subgroup.index H * Subgroup.index K = Subgroup.index H :=
sorry
    \end{lstlisting}\\
    \mycdashline{1-2}
    Aspect & Mathematical Consistency\\
    \mycdashline{1-2}
    Human & Judgment: True\newline
    Explanation: The issue with using `Subgroup.index H' on the right-hand side indicates that the formalization does not fully capture the logic that $H$ is a subgroup of $K$, and that the index should be taken relative to $K$. This is more of a logical modeling issue than a purely mathematical one.\\
    \mycdashline{1-2}
    GPT-4.1 &
    Explanation: The formalization does not accurately represent the mathematical objects and operations present in the natural language statement. The natural language statement involves three subgroups $H \leq K \leq G$ and asserts the equality $|G:H| = |G:K| \cdot |K:H|$, where $|G:H|$ denotes the index of $H$ in $G$, $|G:K|$ the index of $K$ in $G$, and $|K:H|$ the index of $H$ in $K$. In the Lean code, the theorem is stated as `Subgroup.index H * Subgroup.index K = Subgroup.index H`, which is incorrect: it should be `Subgroup.index K * Subgroup.index H = Subgroup.index H` if following the code's order, but even then, the left-hand side should be `Subgroup.index K * Subgroup.index H` only if the indices are taken in the correct groups. More importantly, the code uses `Subgroup.index H`, which is the index of $H$ in $G$, and `Subgroup.index K`, which is the index of $K$ in $G$, but it does not represent $|K:H|$ (the index of $H$ in $K$). There is no mention or calculation of the index of $H$ in $K$ in the code. Therefore, the code does not accurately formalize the mathematical objects and operations (specifically, the indices involved) as described in the natural language statement.\newline
    Judgment: False\\
    \bottomrule\\
  \end{tabular}
  \caption{Example 3 of incorrect LLM judgment in Lean4.}
  \label{tab:case_3}
\end{table*}

\begin{table*}[t]
  \centering
  \small
  \begin{tabular}{p{0.14\textwidth} p{0.8\textwidth}}
    \toprule
    NL Statement & Prove that for every natural number $n$, and for every real number $x \neq \frac{k\pi}{2^t}$ ($t=0,1, \dots, n$; $k$ any integer) $ \frac{1}{\sin{2x}}+\frac{1}{\sin{4x}}+\dots+\frac{1}{\sin{2^nx}}=\cot{x}-\cot{2^nx}  $\\
    \mycdashline{1-2}
    GPT-4.1\newline Formalization &
    \begin{lstlisting}[language=isabelle,aboveskip=-7pt,belowskip=-12pt]
theorem sum_reciprocal_sin_eq_cot_diff
  (n : ℕ) (x : ℝ)
  (hx : ∀ (t : ℕ), t ≤ n → ∀ (k : ℤ), x ≠ (k * Real.pi) / (2^t : ℝ)) :
  (∑ i in Finset.range n.succ, 1 / Real.sin (2^(i+1) * x))
  = Real.cot x - Real.cot (2^n * x) :=
sorry
    \end{lstlisting}\\
    \mycdashline{1-2}
    Aspect & Mathematical Consistency\\
    \mycdashline{1-2}
    Human & Judgment: False\newline
    Explanation: `Finset.range n.succ' gives the set ${0, 1, \dots, n}$; however, the summation should be over ${0, 1, \dots, n - 1}$ when the summand is $\frac{1}{\sin(2^{i+1}x)}$. This indicates a mathematical inconsistency in the indexing.\\
    \mycdashline{1-2}
    GPT-4.1 &
    Explanation: The formalized Lean4 code accurately represents the mathematical objects and operations in the natural language statement. Specifically:
    -The quantification over $n \in \mathbb{N}$ and $x \in \mathbb{R}$ is captured.
    - The condition $x \neq \frac{k\pi}{2^t}$ for $t = 0, 1, ..., n$ and any integer $k$ is represented by the hypothesis.
    - The sum $\frac{1}{\sin{2x}}+\frac{1}{\sin{4x}}+\dots+\frac{1}{\sin{2^n x}}$ is formalized correctly iterates $i$ from $0$ to $n$ and computes $1/\sin(2^{i+1} x)$.
    - The right-hand side $\cot{x} - \cot{2^n x}$ is formalized.
    All mathematical objects (natural numbers, real numbers, integers, sine, cotangent, summation) and their operations are correctly represented in the code.\newline
    Judgment: True\\
    \bottomrule\\
  \end{tabular}
  \caption{Example 4 of incorrect LLM judgment in Lean4.}
  \label{tab:case_4}
\end{table*}

\begin{table*}[t]
    \centering
    \small
    \begin{tabular}{p{0.16\textwidth} p{0.78\textwidth}}
        \toprule
        Purpose & Content\\
        \midrule
        Basic & You are an expert in formal language \{formal\_language\}.\newline You will be given a mathematical statement written in natural language and LaTeX symbols.\newline You will also be given a formal code which attempted to describe the given mathematical statement in \{formal\_language\}.\newline Your task is to evaluate a specific aspect of the formal code.\newline The description of the aspect is: \{\{aspect\_description\}\}\newline Your need to give two things about your evaluation:\newline 1. the judgement of whether the formal code satisfies this aspect. This should be a binary value in "True" or "False".\newline 2. the detailed explanation of your judgement.\newline You should wrap your final results in a way illustrated as the following:\newline \%\%\%\%\%\%\%\%\%\%\newline Explanation: Your Detailed Explanation\newline Judgement: Your Binary Judgement\newline\%\%\%\%\%\%\%\%\%\%\newline Strictly follow the instructions that have been claimed.\\
        \midrule
        Overall & Is the formalized code a correct formalization of the natural language statement?\\
        \midrule
        Logical Preservation & Does the formalized code capture the logical structure and content of the natural language statement?\\
        \midrule
        Mathematical Consistency & Does the formalized code accurately represent mathematical objects and operations present in the natural language statement?\\
        \midrule
        Formal Quality & Does the formalized code demonstrate high quality in terms of structural clarity and usefulness?\\
        \midrule
        Pre-arg Structure & Does the formalized code reflect the \emph{inherent predicate-argument structure} of the natural language statement?\\
        \midrule
        Quantification & Does the formalized code accurately formalize all \emph{quantifiers}, such as universal and existential, present in the natural language statement?\\
        \midrule
        Formula & Are all \emph{mathematical formulas and expressions} in the natural language statement, such as equations and inequalities, correctly and completely represented in the formalized code?\\
        \midrule
        Relation & Are the \emph{logical and mathematical relationships between propositions} in the natural language statement preserved in the formalized code?\\
        \midrule
        Concept & Are all \emph{mathematical concepts} mentioned in the natural language statement, such as integers, fractions, real numbers, complex numbers, derivatives, integrals, vectors, matrices, probabilities, expectations, and variances, correctly formalized in the formalized code?\\
        \midrule
        Constant & Are all \emph{mathematical constants} mentioned in the natural language statement, such as 1, $\frac{2}{3}$, $\pi$, $e$, properly included in the formalized code?\\
        \midrule
        Operator & Are all \emph{mathematical operators} used in the natural language statement, such as addition, subtraction, multiplication, division, summation, exponentiation, and product, correctly represented in the formalized code?\\
        \midrule
        Conciseness & Is the formalized code expressed in a minimal, non-redundant form, avoiding unnecessary repetition or complexity?\\
        \midrule
        Logical Consistency & Is the formalized code \emph{internally coherent and contains no contradictions} under the logical rules of the relevant formal system?\\
        \bottomrule\\
    \end{tabular}
    \caption{LLM-as-a-judge prompts}
    \label{tab:prompt}
\end{table*}

\end{document}

%% file: isabelle.tex
\usepackage{listings}
\usepackage{setspace}
\usepackage{color}

\definecolor{isarblue}{HTML}{006699}
\definecolor{isarfaintblue}{rgb}{0.0, 0.75, 1.0}
\definecolor{isargreen}{HTML}{009966}
\definecolor{red}{HTML}{990000}
\definecolor{patriarch}{rgb}{0.5, 0.0, 0.5}

\lstdefinelanguage{isabelle}{%
    keywords=[1]{theory,type_synonym,datatype,fun,abbreviation,definition,proof,lemma,theorem,qed,corollary,have,hence,also,finally,ultimately,moreover,using,\{},
    keywordstyle=[1]\bfseries\color{isarblue},
    keywords=[2]{where,assumes,shows,fixes,and,begin,end,imports},
    keywordstyle=[2]\bfseries\color{isargreen},
    keywords=[3]{if,then,else,case,SOME,let,in,O},
    keywordstyle=[3]\color{isarblue},
    keywords=[4]{ATP},
    keywordstyle=[4]\it\color{patriarch},
    keywords=[5]{show,assume,obtain},
    keywordstyle=[5]\bfseries\color{isarfaintblue},
    keywords=[6]{<proof>},
    keywordstyle=[6]\color{yellow},
}

\lstdefinestyle{isabelle}{%
  language=isabelle,
  escapeinside={&}{&},
  columns=fixed,
  extendedchars,
  basewidth={0.5em,0.45em},
  basicstyle=\singlespacing\ttfamily\small,
  mathescape,
  morecomment=[s][\bfseries\color{red}]{(*}{*)},
  morecomment=[l][\bfseries]{####},
}